\documentclass[acmtog, nonacm]{acmart}

\newenvironment{datamaterial}%
{ \vspace{-0.15cm}%
    \small\noindent{\bfseries Availability of Data and Material:}\par%
    \noindent\ignorespaces}%
{ \par\noindent%
\ignorespacesafterend }%

\let\oldmaketitle\maketitle
\renewcommand{\maketitle}{%
  \oldmaketitle%
  \thispagestyle{plain}%
  \pagestyle{plain}}

\newcommand{\ifanonymize}[2]{#2}

\usepackage{caption}
\usepackage{subcaption}
\usepackage[printonlyused]{acronym}
\usepackage{natbib}
\usepackage{hyphenat}

\usepackage{siunitx}
\usepackage{booktabs}
\usepackage{placeins}  

\usepackage{pgfcalendar}
\newcount\myjuliandate
\newcount\myjuliantoday
\newcommand{\DaysTo}[3]{%
\pgfcalendardatetojulian{\year-\month-\day}{\myjuliantoday}%
\pgfcalendardatetojulian{#1-#2-#3}{\myjuliandate}%
\advance\myjuliandate by-\myjuliantoday\relax
\the\myjuliandate
}

\begin{document}

\title{A Benchmark for Electrical Load Forecasting Across Grid Levels: Time-Series Transformers Outperform Established Methods}

\author{Matthias Hertel}
\authornote{Matthias Hertel and Sebastian Pütz contributed equally.}
\email{matthias.hertel@kit.edu}
\author{Sebastian Pütz}
\authornotemark[1]
\authornotemark[2]
\email{sebastian.puetz@kit.edu}
\author{Jonathan Kolar}
\email{jonathan.kolar@student.kit.edu}
\author{Benjamin Schäfer}
\authornotemark[2]
\email{benjamin.schaefer@kit.edu}
\author{Ralf Mikut}
\email{ralf.mikut@kit.edu}
\author{Veit Hagenmeyer}
\email{veit.hagenmeyer@kit.edu}
\affiliation{%
  \institution{Karlsruhe Institute of Technology, Germany; \textsuperscript{\textdagger}Helmholtz AI}
  \country{Germany}
}

\renewcommand{\shortauthors}{M. Hertel, S. Pütz, J. Kolar, B. Schäfer, R. Mikut, V. Hagenmeyer}

\begin{abstract}
Accurate load forecasting at multiple grid levels is essential for future smart grids, ranging from aggregated control area forecasts for balancing supply and demand to forecasts of individual end-consumer loads for demand-side management and energy management systems.
We present a comprehensive benchmark for load forecasting across grid levels, comprising three datasets that represent a transmission system operator control area, low-voltage grid feeders, and individual end consumers.
We evaluate ten methods for short-term load forecasting and find that Transformer-based approaches consistently outperform established methods, reducing forecast error by 6.6--10.7\,\%. To analyze the impact of architectural design, we introduce YAformer, a flexible Transformer architecture that integrates modifications from prior work and is optimized via hyperparameter optimization. However, the standard Transformer achieves superior performance, suggesting that these architectural modifications are not required for accurate load forecasting.
We further evaluate the Transformer-based time-series foundation model Chronos-2, which demonstrates competitive zero-shot performance on two datasets but fails to accurately capture special events in the TSO data.
Detailed analyses reveal model-specific strengths and weaknesses, and ablation studies highlight the importance of long input contexts, covariates and continuous retraining -- aspects that are often overlooked in the time-series forecasting literature.
\end{abstract}



\received{24 April 2026}

\maketitle

\begin{datamaterial}
All three datasets come from public sources. Our code and our preprocessed versions of the datasets are made available at \url{https://github.com/KIT-IAI/load-forecasting-benchmark}. 
\end{datamaterial}

\begin{acronym}[LightGBM]
\acro{ARIMA}{Autoregressive Integrated Moving Average}
\acro{CNN}{convolutional neural network}
\acro{DM test}{Diebold--Mariano test}
\acro{EHV}{extra-high voltage}
\acro{GAM}{Generalized Additive Model}
\acro{GEFCom}{Global Energy Forecasting Competition}
\acro{HPC}{high-performance computing}
\acro{HPO}{hyperparameter optimization}
\acro{HV}{high voltage}
\acro{kNN}{k-Nearest Neighbor}
\acro{LightGBM}{Light Gradient Boosting Machine}
\acro{LSTM}{long short-term memory}
\acro{LV}{low voltage}
\acro{MAE}{mean absolute error}
\acro{MLP}{multilayer perceptron}
\acro{nMAE}{normalized \acs{MAE}}
\acro{nRMSE}{normalized \acs{RMSE}}
\acro{ReLU}{rectified linear unit}
\acro{RMSE}{root mean square error}
\acro{RQ}{research question}
\acro{TFT}{Temporal Fusion Transformer}
\acro{TSFM}{Time-Series Foundation Model}
\acro{TSO}{transmission system operator}
\acro{YAformer}{yet another Transformer}
\end{acronym}

\section{Introduction}

The transition toward a climate-neutral energy system affects electricity demand and supply. Electrification of the heat, transport, and industrial sectors is increasing overall demand, while the growing integration of decentralized renewable generation at medium- and low-voltage levels introduces greater variability. As a result, maintaining the real-time balance between supply and demand becomes more challenging, and existing grid infrastructure reaches its limits during periods of peak load or high generation~\ifanonymize{\cite{cakmak_using_2022_anonymized}}{\cite{cakmak_using_2022}}. In this context, smart grid technologies aim to improve the utilization of existing assets~\cite{ramchurn_putting_2012}, but their effectiveness relies on accurate electrical load forecasting across multiple levels of the energy system. At the \ac{EHV} and the \ac{HV} level, forecasts support dispatch and outage planning as well as balancing energy procurement; at the \ac{LV} level, they enable congestion management, demand response, and storage scheduling; and at the client level, they are essential for energy management and peak shaving. These applications are characterized by differing data properties, ranging from relatively stable and aggregated time series at the \ac{EHV} and the \ac{HV} level to highly heterogeneous and volatile patterns at the low-voltage and client levels, often with abrupt changes due to newly installed assets. At the same time, the ongoing rollout of measurement infrastructure, such as smart meters, is increasing data availability, creating new opportunities for data-driven forecasting approaches. The integration of covariates, such as weather data and calendar information, is crucial to achieve accurate load forecasts~\cite{haben_core_2023}.


Recent publications on time-series forecasting make use of the Transformer model~\cite{vaswani_attention_2017}, including Transformer variants particularly designed for time-series forecasting, such as LogSparse Transformer~\cite{li_enhancing_2019}, \ac{TFT}~\cite{lim_temporal_2021}, Informer~\cite{zhou_informer_2021}, FEDformer~\cite{zhou_fedformer_2022}, Autoformer~\cite{wu_autoformer_2021} and PatchTST~\cite{nie_time_2023}.
Since many different time-series Transformer models exist, the present paper analyzes which of the architectural modifications proposed in the literature are beneficial for electrical load forecasting, in order to derive recommendations for practical use cases.

\acp{TSFM}~\cite{liang_foundation_2024} are a new paradigm in time-series forecasting. \acp{TSFM} are pretrained on a variety of time-series datasets and applicable to new time series in a zero-shot setting without the need for additional training.
Most \acp{TSFM} are built on Transformers, underlining the need for analyzing the architectural design of time-series Transformers.

When new forecasting methods are proposed, such as the time-series Transformers and \acp{TSFM}, it is important to benchmark them against established methods to evaluate whether they improve the model accuracy over the state of the art~\cite{hewamalage_forecast_2023}.
Because data characteristics vary across load aggregation levels, the present paper benchmarks Transformers and \acp{TSFM} against established methods on three different datasets, representing the \ac{TSO} level, the \ac{LV} feeder level, and the client level.



\noindent
In this context, our contributions are as follows:
\begin{enumerate}
    \item We select and process three highly relevant and representative datasets for benchmarking of newly developed load forecasting methods across grid levels, representing a \ac{TSO} control area, \ac{LV} grid feeders and individual end consumers.
    \item We study the effect of Transformer architectures on the forecast accuracy by introducing \acs{YAformer}, a flexible Transformer architecture that integrates multiple modifications from architectures proposed in the literature and is optimized with a \ac{HPO}.
    \item We compare standard Transformers, \acs{YAformer} and the \ac{TSFM} Chronos-2~\cite{ansari_chronos-2_2025} with seven established forecasting methods and one baseline and highlight their advantages and disadvantages in terms of forecast accuracy and runtime.
\end{enumerate}

The remainder of the paper is structured as follows:
Section~\ref{sec:related-work} presents the related work on time-series Transformers, \acp{TSFM} and benchmarking.
Section~\ref{sec:methodology} introduces \acs{YAformer} and the comparison methods.
Section~\ref{sec:experimental-setup} presents the datasets and experimental setup.
Section~\ref{sec:results} describes the results, which are discussed in Section~\ref{sec:discussion}.
Section~\ref{sec:conclusion} concludes the paper and highlights potential future work.

\section{Related Work}
\label{sec:related-work}

Electrical load forecasting has a long history, with first approaches dating back to the 1940s~\cite{hong_energy_2014}.
Since then, it became an active field of research, with recently more than 1,000 publications published annually\footnote{A SCOPUS search for publications with "load forecast*" in the title returned 1,398 documents published in the year 2025 alone.}.
Despite the large number of publications in the field, literature reviews on load forecasting largely agree that there is a small set of established methods that are frequently used in research and in practice~\cite{hong_energy_2020, scheidt_data_2020, haben_review_2021}. This includes statistical methods like \ac{ARIMA}, machine learning methods like \ac{kNN} and gradient boosting machines, and deep learning methods like \acp{MLP}, \acp{CNN} and \ac{LSTM} networks.

The Transformer is a deep learning model developed for machine translation~\cite{vaswani_attention_2017} that led to improvements in many domains, including language modeling~\cite{radford_language_2019}, computer vision~\cite{dosovitskiy_image_2020} and tabular data~\cite{hollmann_tabpfn_2023}.
The Transformer was also adopted for time-series forecasting~\cite{wu_deep_2020} and electrical load forecasting in particular~\cite{zhang_short-term_2022}.
Many Transformer architectures for time-series forecasting exist, including LogSparse Transformer~\cite{li_enhancing_2019}, \ac{TFT}~\cite{lim_temporal_2021}, Informer~\cite{zhou_informer_2021}, FEDformer~\cite{zhou_fedformer_2022}, Autoformer~\cite{wu_autoformer_2021} and PatchTST~\cite{nie_time_2023}.
The architectural modifications of these models include \ac{LSTM} layers~\cite{lim_temporal_2021}, LogSparse attention~\cite{li_enhancing_2019}, convolutional attention~\cite{li_enhancing_2019}, ProbSparse attention~\cite{zhou_informer_2021}, attention based on autocorrelation~\cite{wu_autoformer_2021}, max pooling~\cite{zhou_informer_2021} (called "self-attention distilling" in~\cite{zhou_informer_2021}), time series decomposition~\cite{zhou_fedformer_2022} and patched inputs~\cite{nie_time_2023, ansari_chronos-2_2025}.
However, the evaluations in these publications do not generalize to electrical load forecasting, as they are often performed on datasets from different domains, focusing on long forecast horizons, and neglecting exogenous features which are important for electrical load forecasting~\cite{haben_review_2021}.
While works on electrical load forecasting with Transformers show their potential~\ifanonymize{\cite{zhang_short-term_2022, hertel_transformer_2023_anonymized, giacomazzi_short-term_2023}}{\cite{zhang_short-term_2022, hertel_transformer_2023, giacomazzi_short-term_2023}}, an evaluation of multiple Transformer architectures on electrical load datasets representing different aggregation levels is currently missing.


Recently, \acp{TSFM} were introduced, which are generalized time-series forecasting models trained on large amounts of time-series data from different domains, often including synthetic data, thereby being applicable to new time series in a zero-shot setting (i.e., without being trained on data from the particular time series).
The \acp{TSFM} Moirai~\cite{woo_unified_2024}, Moirai-2~\cite{liu_moirai_2026}, TimeGPT~\cite{garza_timegpt-1_2023}, Lag-Llama~\cite{rasul_lag-llama_2023}, Chronos~\cite{ansari_chronos_2024} and Chronos-2~\cite{ansari_chronos-2_2025} are based on Transformers. While time-series Transformers are predominantly used for \acp{TSFM}, other architectures also exist, including TabPFN-TS~\cite{hoo_tables_2025}, which is based on a tabular Transformer model, and TiRex~\cite{auer_tirex_2025}, which is based on an extended \ac{LSTM}. We select Chronos-2 for the experiments, because it handles covariates well and outperforms other \acp{TSFM} on several electrical load forecasting datasets~\ifanonymize{\cite{kreusel_covariates_2026_anonymized, hertel_explainable_2026_anonymized, cheong_assessing_2026}}{\cite{kreusel_covariates_2026, hertel_explainable_2026, cheong_assessing_2026}}.

With the development of \acp{TSFM}, large benchmark datasets for the evaluation of forecasting models were introduced, but the ability to generalize the results to electrical load forecasting is limited.
The Monash Time Series Forecasting Archive~\cite{godahewa_monash_2021} contains 25 time series datasets, including two electrical load datasets, but it does not incorporate covariates.
BuildingsBench~\cite{emami_buildingsbench_2023} contains synthetic and real load time series from buildings, including calendar features but no weather data.
Fev-bench~\cite{shchur_fev-bench_2025} comprises 100 forecasting tasks, including electrical load data from ENTSO-E and from the \acp{GEFCom} \cite{hong_energy_2020}. \citet{shchur_fev-bench_2025} compare multiple \acp{TSFM} on fev-bench and find that Chronos-2~\cite{ansari_chronos-2_2025} performs best, but no comparisons are made to models trained from scratch.
GIFT-Eval~\cite{aksu_gift-eval_2024} contains only one electrical load dataset, which does not incorporate covariates.
Therefore, the present paper evaluates Chronos-2 and models trained from scratch on a variety of electrical load datasets representing different aggregation levels.

Complementary work evaluates the usage of TSFMs for peak load forecasting in the \ac{LV} grid~\cite{kaas_probabilistic_2026}.
Their work focuses on probabilistic forecasting and specialized metrics for peak forecasting, whereas our work focuses on deterministic forecasting, experiments across grid levels, and evaluations of overall forecast accuracy.


\section{Methodology}
\label{sec:methodology}

This section introduces a flexible Transformer architecture, the comparison methods, and the hyperparameter optimization.\footnote{Code is available on GitHub: \url{https://github.com/KIT-IAI/load-forecasting-benchmark}}

\subsection{\acs{YAformer}: a flexible Transformer architecture}

\begin{figure}
    \centering
    \includegraphics[width=\linewidth]{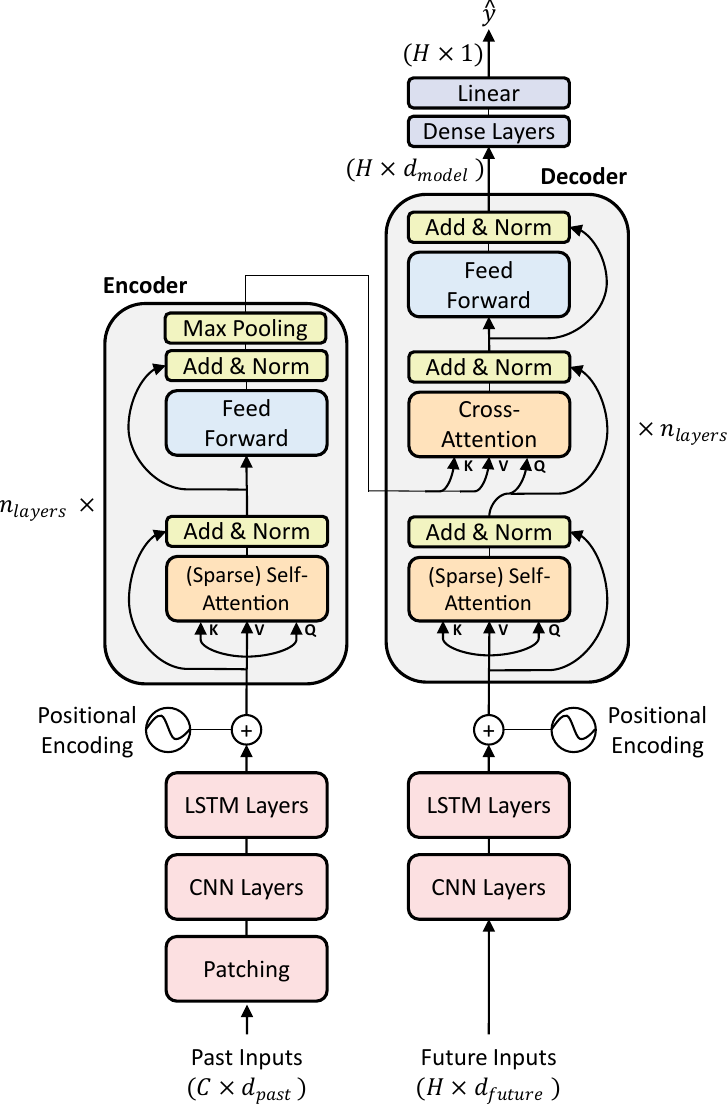}
    \caption{Overview of the YAformer architecture based on an encoder-decoder Transformer. The inputs go through patching, CNN and LSTM layers. Sparse self-attention and max pooling can be optionally added in the encoder and decoder layers. The last linear layer produces a point forecast for the next $H$ time steps. Figure based on~\cite{vaswani_attention_2017}.}
    \label{fig:yaformer-architecture}
\end{figure}

We introduce \ac{YAformer}, a flexible architecture that integrates architectural modifications to the standard Transformer proposed in prior work. Its design is controlled by a set of hyperparameters, allowing to represent a variety of time-series Transformer variants. This allows us to systematically compare an optimized YAformer model with the standard Transformer, thereby quantifying the impact of architectural modifications on forecasting accuracy.
An overview of the model architecture is given in Figure~\ref{fig:yaformer-architecture}.
The model gets the past load, past covariates and future covariates as input.
The basis of \ac{YAformer} is the standard encoder-decoder Transformer~\cite{vaswani_attention_2017} adapted for time series.
For context length $C$, the encoder gets $C$ vectors as input, each representing the load and covariates (weather data and calendar information) of one past time step.
These vectors have one entry for the load and one for each covariate, so their size $d_\mathrm{past}$ equals the number of covariates plus one.
The vectors go through a linear layer to embed them in the hidden dimension $d_\mathrm{model}$.
A stack of $n_\mathrm{layers}$ Transformer encoder layers creates context-aware embeddings, using self-attention with keys $K$, values $V$ and queries $Q$ to retrieve information from the embedded input series.
The encoder output has dimension $C \times d_\mathrm{model}$.
The decoder gets $H$ vectors as input for a forecast horizon $H$, each representing the covariates of one future time step, so their size $d_\mathrm{future}$ equals the number of covariates.
Like in the encoder, these vectors are embedded to dimension $d_\mathrm{model}$ via a linear layer.
A stack of $n_\mathrm{layers}$ Transformer decoder layers create context-aware embeddings, using self-attention to retrieve information from the output of the previous layer and cross-attention to retrieve information about the past time steps from the encoder output.
The decoder output has dimension $H \times d_\mathrm{model}$ and goes through a stack of dense layers with \ac{ReLU} activation and a final linear output layer with one unit to create a point forecast of dimension $H \times 1$, containing the forecast values for the next $H$ time steps.

The following architectural modifications can be added via hyperparameters:
\begin{itemize}
    \item Patching: Neighboring past input vectors are concatenated before going through the linear embedding layer. The patch size $p$ determines the number of vectors that are concatenated.
    Since $p$ vectors get concatenated, the encoder input is transformed from dimension $C \times d_\mathrm{past}$ to $\dfrac{C}{p} \times (d_\mathrm{past} \cdot p)$.
    \item \Ac{CNN} layers: To encode local information, the linear embedding layer is replaced by a stack of convolutional layers with $d_\mathrm{model}$ filters. The number of convolutional layers is controlled by a hyperparameter $n_\mathrm{conv}$ and the kernel width by a hyperparameter $w_\mathrm{conv}$.
    \item \Ac{LSTM} layers: Before the Transformer layers in the encoder and decoder, a stack of \ac{LSTM} layers processes the embedded vectors sequentially.
    \item Max pooling: After each Transformer encoder layer, max pooling is used to reduce the length of the series. The pooling width $w_\mathrm{max}$ is controlled by a hyperparameter.
    \item Sparse attention: Sparse attention restricts the self-attention mechanisms in the encoder and decoder to use information only from the vectors of previous vectors with a logarithmic distance to the embedded position (the vector itself and 1, 2, 4, 8, ... steps before). A hyperparameter is used to switch between full attention and LogSparse attention~\cite{li_enhancing_2019}.
\end{itemize}

\subsection{Comparison methods}

The following methods are selected for comparison, comprising a number of established methods~\cite{hong_energy_2020, scheidt_data_2020, haben_review_2021} as well as recently developed Transformer models~\cite{vaswani_attention_2017, lim_temporal_2021} and the \ac{TSFM} Chronos-2~\cite{ansari_chronos-2_2025}.

\paragraph{Persistence baseline} The baseline predicts the load value from the same hour on the last observed day that has the same type as the predicted day, distinguishing between three types: workday, Saturday, and Sunday/holiday.

\paragraph{Linear model} Like the models from \citet{zeng_are_2023}, the linear model gets a feature vector as input and predicts the next $H$ values. The model consists of a single linear layer with $n_\mathrm{features} \times H$ weights. It is a global model, meaning that the weights are shared for all time series. The feature vector is the concatenation of $C$ past vectors, containing the load and covariates for the previous time steps, and $H$ future vectors, only containing the covariates for the future time steps.

\paragraph{\Acf{LightGBM}} \acs{LightGBM}~\cite{ke_lightgbm_2017} is a machine learning method that builds an ensemble of decision trees using gradient boosting. It is designed to train efficiently and handle large datasets with high performance. We train $H$ individual models that predict the $H$ future values based on past observations and future covariates.

\paragraph{\Acf{MLP}} The \ac{MLP} gets a flat vector as input, which is the concatenation of the $C$ past and $H$ future vectors. Multiple fully-connected layers with \ac{ReLU} activation are stacked on top of each other. The final layer is a linear output layer with $H$ units returning the forecast values.

\paragraph{\Acf{CNN}} The \ac{CNN} gets a sequence of $C + H$ vectors as input. The load value in the future vectors is set to zero as it is not known at prediction time. Multiple convolutional layers with a stride of one are stacked on top of each other. The number of filters and the filter width are controlled via hyperparameters. The output of the last convolutional layer is flattened and fed into a stack of fully-connected layers with \ac{ReLU} activation, where the last layer has $H$ units and linear activation to create the forecast values.

\paragraph{\Acf{LSTM}} A hyperparameter for the \ac{LSTM} allows to switch between an encoder\hyp{}decoder model and an encoder-only model. The inputs to the encoder and decoder are the same as with \ac{YAformer}. The vectors get embedded to dimension $d_\mathrm{model}$ by linear layers. The encoder and decoder are bidirectional \acp{LSTM}. The last hidden vector of the encoder is fed to the decoder as its first hidden vector. The decoder output for each future time step goes through a stack of linear layers and a final output layer with one unit to create the $H$ forecast values.
In the encoder-only variant, all input vectors are fed to the encoder \ac{LSTM} stack, whose output is flattened and goes through a stack of linear layers and a final output layer with $H$ units to create the $H$ forecast values.

\paragraph{N-HITS} N-HITS~\cite{challu_nhits_2023} is a stack of \acp{MLP}, each of which returns a time series with a different time resolution. The forecast is finally computed as the sum of the returned (interpolated) time series.

\paragraph{Transformer} We use a standard encoder-decoder Transformer~\cite{vaswani_attention_2017}. It gets the same inputs as \ac{YAformer}, but does not make use of the architectural modifications introduced for \ac{YAformer}.

\paragraph{\Acf{TFT}}
\Ac{TFT}~\cite{lim_temporal_2021} is an encoder\hyp{}decoder model, receiving the same inputs as \ac{YAformer}. The first layer of the encoder and decoder is a variable selection layer. It is followed by unidirectional LSTM layers and a single Transformer layer. Causal masking is used in the attention mechanism to prevent the Transformer layer from using information of time steps after the prediction time step.

\paragraph{Chronos-2} Chronos-2~\cite{ansari_chronos-2_2025} is a \ac{TSFM} based on a Transformer with patched inputs. It is pretrained on a large dataset containing real time series data from various domains, as well as synthetic data with covariate dependencies. It is the best open model on the GIFT-Eval benchmark, particularly outperforming the \acp{TSFM} Moirai~\cite{woo_unified_2024}, TimesFM~\cite{das_decoder-only_2024} and TabPFN-TS~\cite{hoo_tables_2025}.

\subsection{\Acf{HPO}}

The architecture modifications of \ac{YAformer} can be controlled via model hyperparameters. Note that the standard Transformer is a special case of \ac{YAformer} with patch size 1, zero CNN and LSTM layers, pooling width 1 and full attention.
The hyperparameters of all models (YAformer and the comparison models) are optimized using Bayesian Optimization in weights and biases~\cite{biewald_experiment_2020} with 100 hyperparameter configurations.
All deep learning models are trained with AdamW~\cite{loshchilov_decoupled_2019} and the mean squared error loss function, one validation per epoch on TransnetBW and ten validations per epoch on the other datasets, using early stopping with a patience of ten validations without improvement.
The best hyperparameters found with the \ac{HPO} are given in Appendix~\ref{sec:hyperparameters}.

\begin{table*}[htb]
    \caption{Dataset characteristics.}
    \label{tab:datasets}
    \centering
    \begin{tabular}{lccc}
    \toprule
     \textbf{Dataset} & \textbf{Aggregation} & \textbf{Time series} & \textbf{Period} \\
    \midrule
     TransnetBW & \ac{TSO} control area & 1 & 1.1.2015--31.12.2025 \\
     FeederBW & \ac{LV} feeders & 200 & 1.4.2023--31.3.2025 \\
     Electricity-287 & Clients & 287 & 1.1.2011--31.12.2014 \\
    \bottomrule
    \end{tabular}
\end{table*}

\section{Experimental Setup}
\label{sec:experimental-setup}

This section describes the three datasets, the task definition and metrics, and the rolling evaluation scheme used in the experiments.

\subsection{Datasets}

The proposed benchmark contains three datasets from different load aggregation levels, thereby representing use cases in the \ac{EHV} and \ac{HV} grid, the low-voltage grid and for end consumers. These datasets are described next. An overview of the datasets is given in Table~\ref{tab:datasets}.\footnote{All three datasets come from public sources and our preprocessed versions are made available to foster future comparisons to our results: \url{https://github.com/KIT-IAI/load-forecasting-benchmark}}

\paragraph{TransnetBW}
TransnetBW is the \ac{TSO} of the German state Baden-Württemberg. We download the load values for the \ac{TSO} control area for the years 2015--2025 from the ENTSO-E transparency platform~\cite{entso-e_transparency_2026}.
We transform the data into hourly resolution by taking the mean of four quarter-hourly power values.
The dataset is enriched with weather data for Baden-Württemberg from the ERA5 reanalysis model, downloaded from Copernicus~\cite{copernicus_climate_change_service_climate_2020}.
We use the air temperature, irradiance, precipitation, wind speed at 10 meters height and a binary holiday feature as covariates.
The year 2023 is used as validation set for the \ac{HPO} and the years 2024 and 2025 are used as test set.

\paragraph{FeederBW}
The FeederBW dataset~\cite{treutlein_real-world_2026} comprises the electrical load from 200 low-voltage feeders in Baden-Württemberg in minutely resolution, together with short-term weather forecasts associated with the post code of each feeder.
The length of the time series is two years, ranging from 1.4.2023 to 31.3.2025.
We transform the load data into hourly resolution and take March 2023 as validation set and the last twelve months (1.4.2024--31.3.2025) as test set.
We use irradiance, air temperature, precipitation, and a binary holiday feature as covariates.

\paragraph{Electricity-287}
To represent the client level, we use public data from the UCI Electrical Load Diagrams\footnote{\url{https://archive.ics.uci.edu/dataset/321/electricityloaddiagrams20112014}}, first mentioned by \citet{rodrigues_load_2018}.
This dataset contains 370 load time series from individual clients of an energy trading company from Portugal.
The data is in 15-minutes resolution, ranging from 1.1.2011--31.12.2014.
A subset of the dataset is contained in the GIFT-Eval benchmark, there called the 'Electricity' dataset.
We take a cleaned version of the original dataset, where time series with lots of missing data or imputation artifacts are removed~\ifanonymize{\cite{hertel_comparison_2025_anonymized}}{\cite{hertel_comparison_2025}}, and reduce it further to time series that have no missing data in the validation or test period.
The resulting dataset has 287 time series, which we transform into hourly resolution.
The original dataset has no weather information, so we combine it with weather data for Portugal from the ERA5 reanalysis model~\cite{copernicus_climate_change_service_climate_2020}.
To the best of our knowledge, we are the first to combine the Electricity dataset with weather data and study the effect of weather covariates on the forecast accuracy.
The same test split as in GIFT-Eval~\cite{aksu_gift-eval_2024} and \mbox{Chronos-2}~\cite{ansari_chronos-2_2025} is used to allow for a fair comparison, meaning that the last 10\,\% (8.8.2014 -- 31.12.2014) are used as test data. We use the same portion of data (15.3.2014 -- 7.8.2014) as validation set.

\subsection{Task definition and metrics}

The task addressed in this work is to forecast the hourly loads for the next four days (96 hours) based on past observations and past and future covariates.
We see this as representative for many applications, which often require planning more than one day ahead.
Forecasts are generated with a stride of one hour, i.e., at every hour, the next 96 values are predicted.
For the datasets with multiple time series (FeederBW and Electricity-287), individual forecasts for every time series are made.
The deep learning models are trained with a global training strategy, which was found to be superior to local and multivariate models~\ifanonymize{\cite{hertel_transformer_2023_anonymized, grabner_global_2023, nie_time_2023}}{\cite{hertel_transformer_2023, grabner_global_2023, nie_time_2023}}, whereas \ac{LightGBM} is trained as local models.

In addition to the weather and holiday covariates, the models trained from scratch get covariates as input that are derived from the timestamp, namely the hour of day, day of week and month (all cyclically encoded as two features).

We use the \ac{MAE} to evaluate the forecasts for the TransnetBW and FeederBW datasets.
To account for the different magnitudes of the time series in the Electricity-287 dataset, the \ac{nMAE} is used instead of the \ac{MAE}.
The \ac{nMAE} is the \ac{MAE} divided by the mean of the time series' absolute ground truth values in the test period.
The metrics are computed per time series on the entire test set and the mean of all time series is reported as the final metric value.
Results for additional metrics are reported in Appendix~\ref{sec:metric-results}.

\subsection{Retraining}

For the datasets with at least one year test period (TransnetBW and FeederBW), new models are trained every month, using the last month as validation data and everything before as training data.
This allows the models to adapt to potential concept drifts.
For the Electricity-287 dataset, a static split into training (1.1.2011--14.3.2014), validation (15.3.2014--7.8.2014) and test data (8.8.2014--31.12.2014) is used.

\subsection{Statistical Comparison}
For statistical evaluation of the observed performance differences, we apply \acp{DM test}~\cite{dieboldComparingPredictiveAccuracy1995} on the per-series loss differentials between competing forecasting models. For each dataset and model, the absolute error per sample is averaged over the forecasting horizon, yielding one aggregated loss value per series and forecast origin. For each pair of models, we compute the corresponding loss differentials. For each series, we separately perform two one-sided \acp{DM test}, resulting in two $p$-values corresponding to the hypotheses that either model~1 or model~2 achieves significantly lower predictive loss. A model is counted as significantly better for a given series if the corresponding one-sided \ac{DM test} rejects the null hypothesis at the $5\%$ significance level. We implement the \ac{DM test} in Python following the implementation of the R forecast package~\cite{hyndmanforecastpackage}.

\section{Results}
\label{sec:results}

 \begin{figure*}
    \centering
    \begin{subfigure}{0.33\textwidth}
        \includegraphics[width=\textwidth]{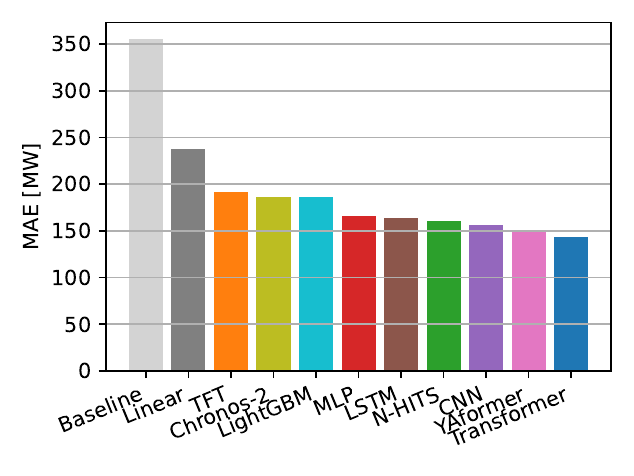}
        \caption{TransnetBW}
        \label{fig:transnetbw-results}
    \end{subfigure}
    \begin{subfigure}{0.33\textwidth}
        \includegraphics[width=\textwidth]{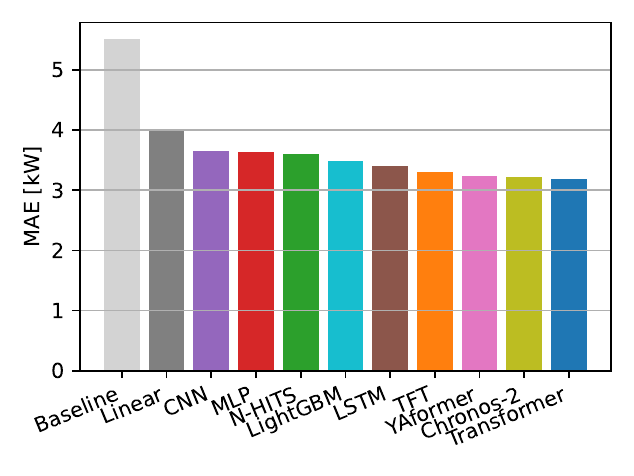}
        \caption{FeederBW}
        \label{fig:feederbw-results}
    \end{subfigure}
    \begin{subfigure}{0.33\textwidth}
        \includegraphics[width=\textwidth]{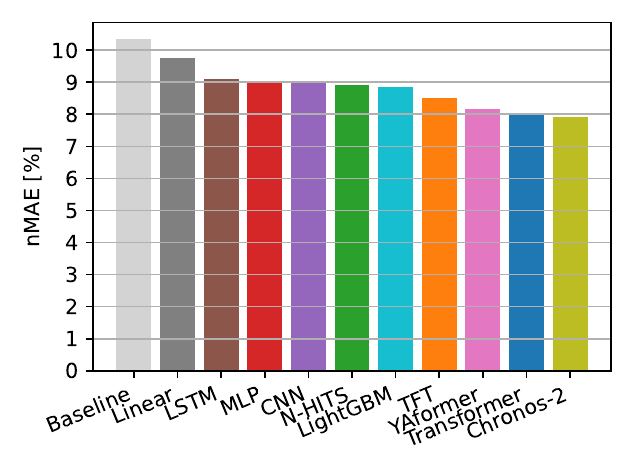}
        \caption{Electricity-287}
        \label{fig:electricity-results}
    \end{subfigure}
    \begin{subfigure}{0.33\textwidth}
        \includegraphics[width=\textwidth]{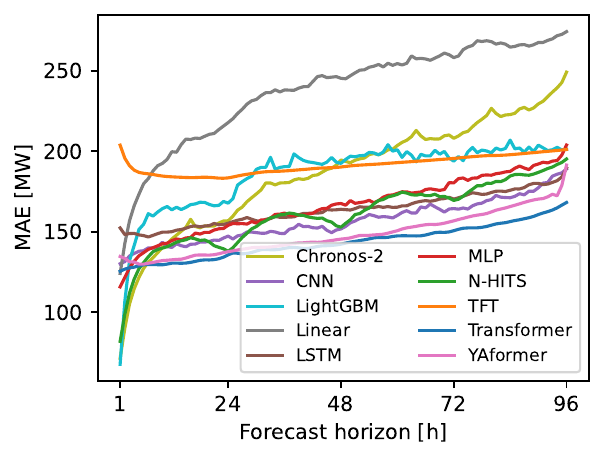}
        \caption{TransnetBW}
        \label{fig:transnetbw-results}
    \end{subfigure}
    \begin{subfigure}{0.33\textwidth}
        \includegraphics[width=\textwidth]{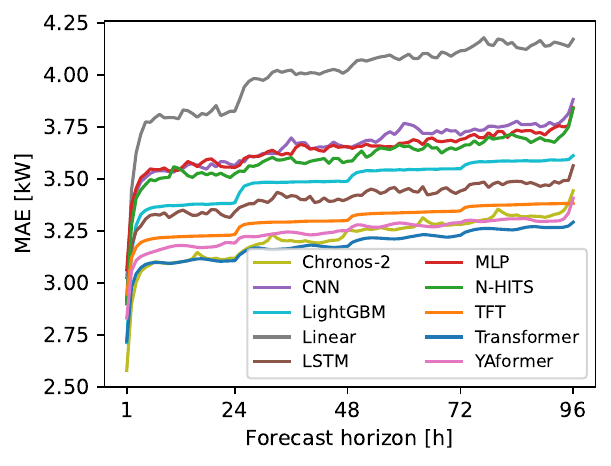}
        \caption{FeederBW}
        \label{fig:feederbw-results}
    \end{subfigure}
    \begin{subfigure}{0.33\textwidth}
        \includegraphics[width=\textwidth]{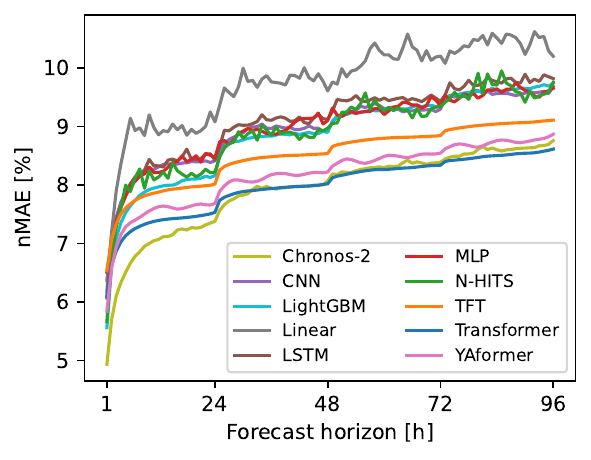}
        \caption{Electricity-287}
        \label{fig:electricity-results}
    \end{subfigure}
    \caption{Benchmarking results on the three datasets.}
    \label{fig:results}
 \end{figure*}

This section describes the forecasting results on the three datasets, as well as runtimes and ablation studies.

\subsection{Model comparison}

We present the model comparison on the three datasets in Figure~\ref{fig:results}.
All models outperform the baseline on all three datasets.
The improvement of the best model compared to the baseline is 59.6\,\% on TransnetBW, 42.3\,\% on FeederBW and 23.3\,\% on Electricity-287, indicating that it is easier to outperform a simple approach on highly aggregated data than on the more volatile \ac{LV} feeder and client-level data.

The Transformer achieves the lowest forecast error on the TransnetBW and FeederBW datasets, whereas it is outperformed by Chronos-2 on the Electricity-287 dataset. Note that the training set of the Electricity-287 dataset is contained in the pretraining data of Chronos-2, so the evaluation on this dataset is not zero-shot.

On all three datasets, the best models are based on time-series Transformer architectures; namely, the two best models on TransnetBW (Transformer and \ac{YAformer}) and the four best models on FeederBW and Electricity-287 (Transformer, Chronos-2, \ac{YAformer} and \ac{TFT}).
The best Transformer model achieves 8.3\,\%, 6.6\,\%, and 10.7\,\% lower forecast errors than the best non-Transformer model (\ac{CNN}, \ac{LSTM} and N-HITS) on TransnetBW, FeederBW and Electricity-287, respectively.
The Transformer outperforms \ac{YAformer} on all three datasets, indicating that the architectural modifications do not improve the forecast error compared to that of a standard Transformer.

The forecast performance of Chronos-2 varies greatly across the three datasets. Chronos-2 is the best model on the Electricity-287 dataset, and second-best on the FeederBW dataset, but has a much larger \ac{MAE} than the best models on the TransnetBW dataset. A closer look at the model performance over the test period in Figure~\ref{fig:transnetbw-origin} reveals that Chronos-2 has large errors around holidays (e.g., 1.1.2024, 28.3.2024, 1.5.2024, 3.10.2024, the Christmas period, 17.4.2025, 3.10.2025).
Figure~\ref{fig:examples-transnetbw} shows four forecasts where Chronos-2 exhibits large errors. The beginning of January is characterized by low loads due to winter holidays, and Chronos-2 fails to adjust the load to normal levels after the holiday period. Chronos-2 predicts a higher load on Good Friday (29.3.2024), potentially because the previous holiday (6.1.2024) did not affect the load because it was on a weekend. Ascension Day (9.5.2024) is predicted correctly, but Chronos-2 does not predict the long weekend. Before Christmas, many people take the full week off, and 24.12.2024 is not an official holiday, so Chronos-2 predicts a load that is too high.
We evaluated Chronos-2 on TransnetBW with additional binary features for long weekends and school holidays, which reduces its \ac{MAE} to 180~MW, but it is still outperformed by the deep learning models trained from scratch.

 \begin{figure*}
     \centering
     \begin{subfigure}{\textwidth}
         \includegraphics[width=\textwidth]{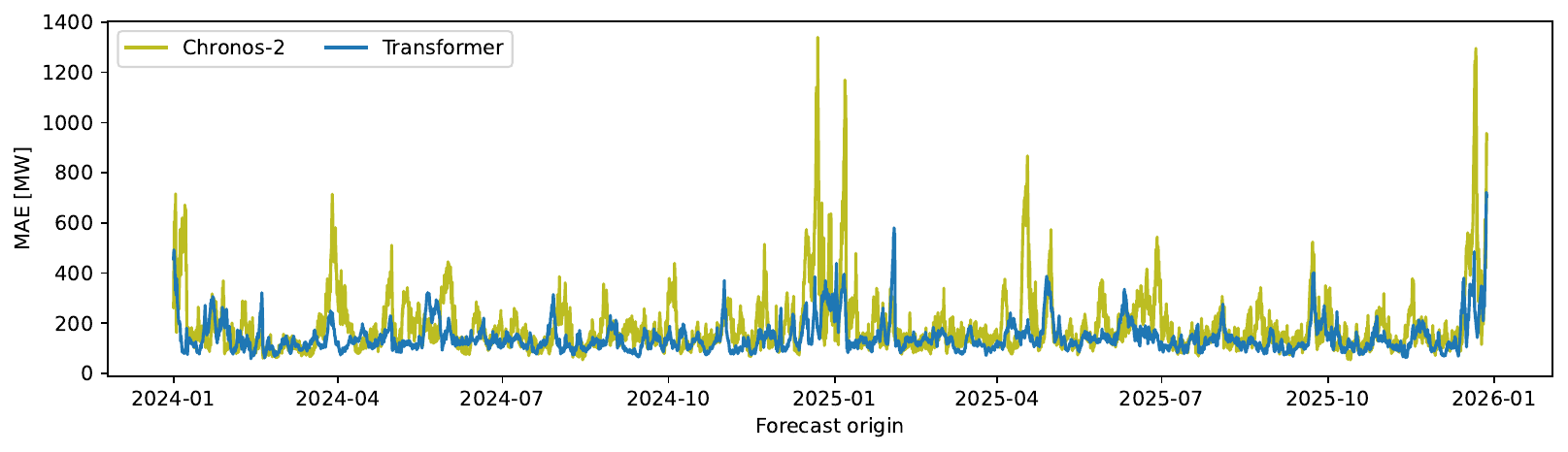}
         \caption{TransnetBW}
         \label{fig:transnetbw-origin}
     \end{subfigure}
     \begin{subfigure}{\textwidth}
         \includegraphics[width=\textwidth]{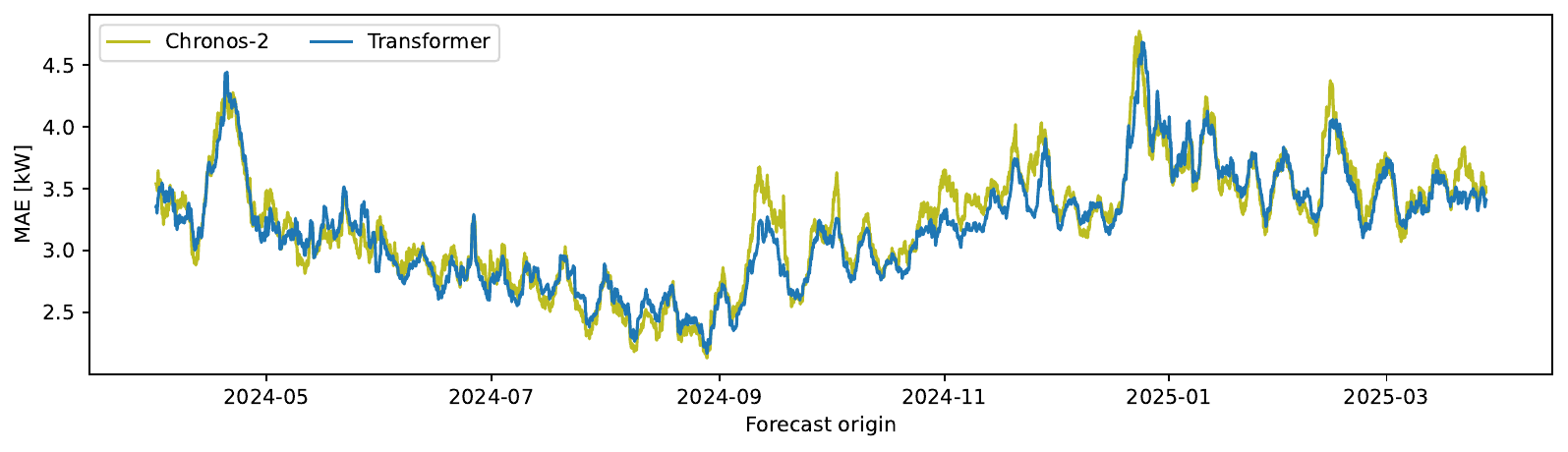}
         \caption{FeederBW}
         \label{fig:feederbw-origin}
     \end{subfigure}
     \begin{subfigure}{\textwidth}
         \includegraphics[width=\textwidth]{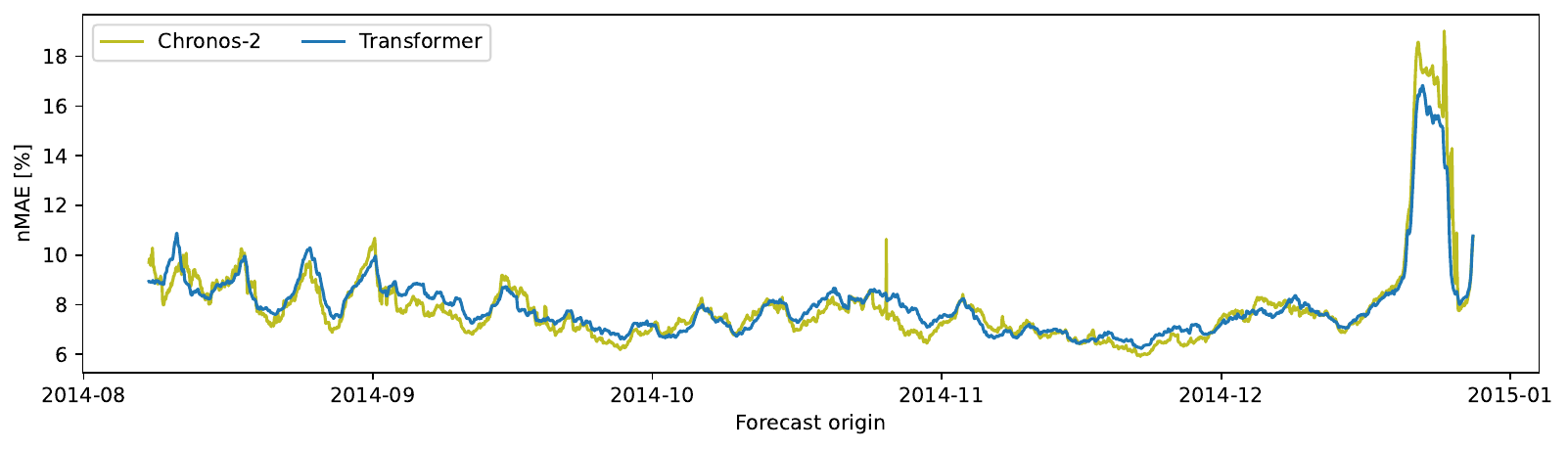}
         \caption{Electricity-287}
         \label{fig:electricity-origin}
     \end{subfigure}
     \caption{Comparison of the forecast error of the Transformer and Chronos-2 over time.}
     \label{fig:results-origin}
 \end{figure*}

On the FeederBW dataset (Figure~\ref{fig:feederbw-origin}), the forecast errors are higher in winter, when the load is generally higher, with spikes in April and around Christmas. In this case, the Transformer outperforms Chronos-2, especially from September to November.
On the Electricity-287 dataset (Figure~\ref{fig:electricity-origin}), Chronos-2 outperforms the Transformer consistently, except for the Christmas period, where both models exhibit large errors.

The forecast error for forecast horizons ranging from one hour to four days are shown in Figure~\ref{fig:results}d-f.
\ac{LightGBM} is the best model for one-step forecasts on the TransnetBW dataset.
Chronos-2 is good for short forecast horizons, outperforming all other models for two to six hours horizon on TransnetBW, for the first six hours on FeederBW, and for the first 31 hours on Electricity-287.
The error of the Transformer increases less with longer horizons, making it the best model for long horizons.


The results of the \acp{DM test} to compare the models per time series are given in Figure~\ref{fig:statistical-tests}. For each pair of models, the number in the figure is the fraction of time series where one model is significantly better than the other model with $p<0.05$.
The TransnetBW dataset contains only one time series, so the fraction of time series can only be 100\% or 0\%. Transformer and \ac{YAformer} are not significantly different, and they outperform all other models significantly.
On the FeederBW dataset, the result is mixed, with the Transformer outperforming Chronos-2 on 46\% of the time series, but vice versa on 20\% of the time series (and no significant difference on the remaining time series).
On the Electricity-287 dataset, Chronos-2 has the best \ac{nMAE}, but it outperforms the Transformer only on 14\% of the time series, whereas the result is vice versa on 39\% of the time series (and no significant difference on the remaining time series). The reason is that Chronos-2 is particularly good on difficult time series with a high \ac{nMAE}, but it is outperformed on many time series with lower \ac{nMAE}.

\begin{figure*}
    \begin{subfigure}{0.49\textwidth}
        \includegraphics[width=\textwidth]{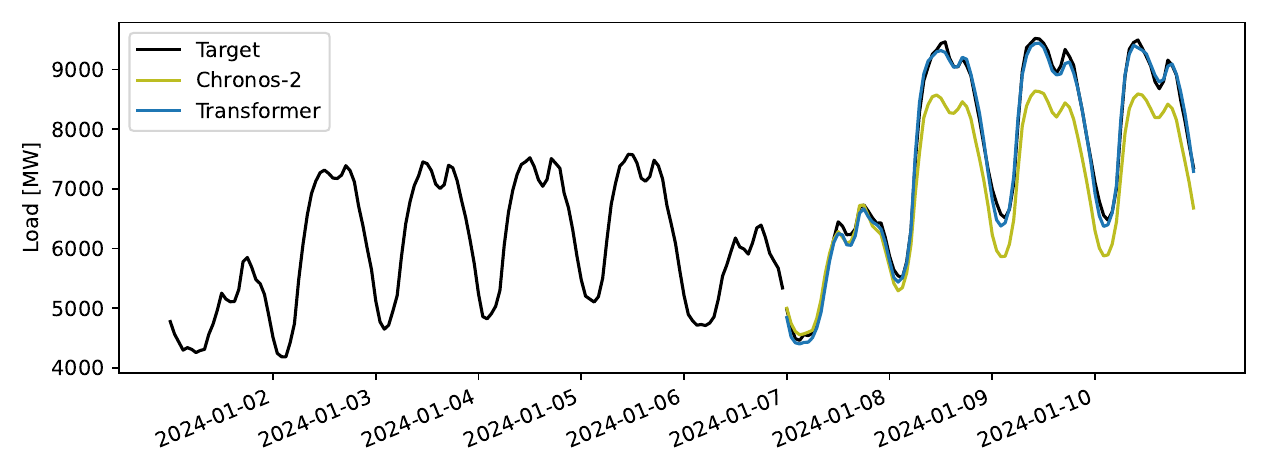}
        \caption{Forecast from 7.1.2024.}
    \end{subfigure}
    \begin{subfigure}{0.49\textwidth}
        \includegraphics[width=\textwidth]{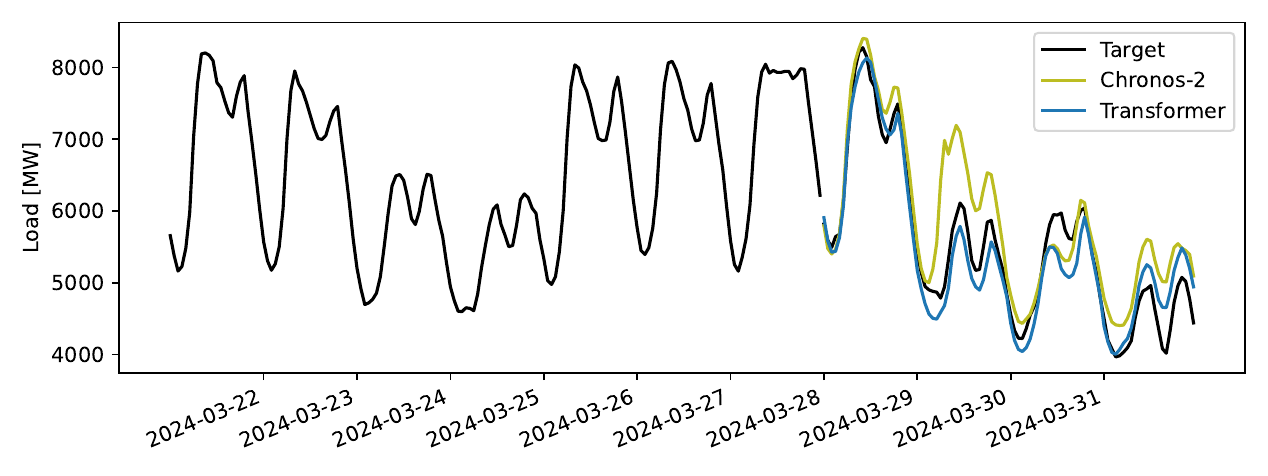}
        \caption{Forecast from 28.3.2024.}
    \end{subfigure}
    \begin{subfigure}{0.49\textwidth}
        \includegraphics[width=\textwidth]{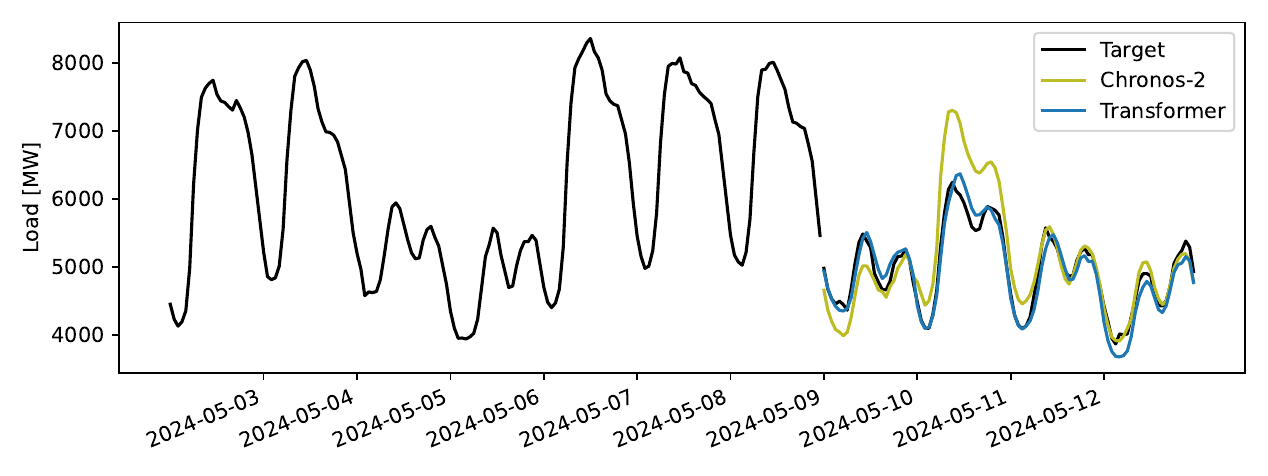}
        \caption{Forecast from 9.5.2024.}
    \end{subfigure}
    \begin{subfigure}{0.49\textwidth}
        \includegraphics[width=\textwidth]{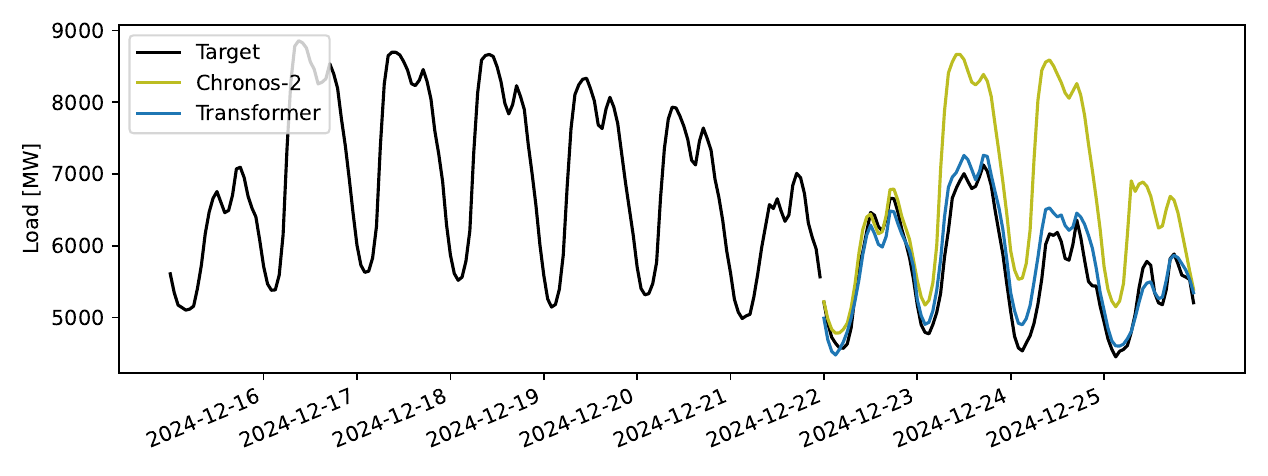}
        \caption{Forecast from 22.12.2024.}
    \end{subfigure}
    \caption{Exemplary forecasts of the Transformer and Chronos-2 on the TransnetBW data, showing cases where Chronos-2 exhibits large errors.}
    \label{fig:examples-transnetbw}
\end{figure*}


\begin{figure*}
    \centering
    \begin{subfigure}{0.33\textwidth}
        \includegraphics[width=\textwidth]{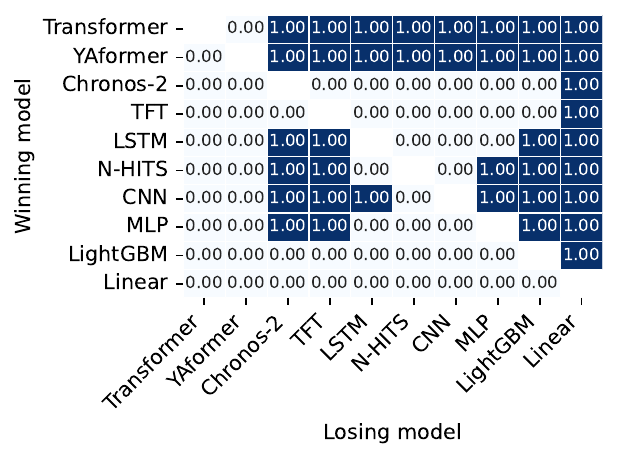}
        \caption{TransnetBW}
        \label{fig:transnetbw-statistical-test}
    \end{subfigure}
    \begin{subfigure}{0.33\textwidth}
        \includegraphics[width=\textwidth]{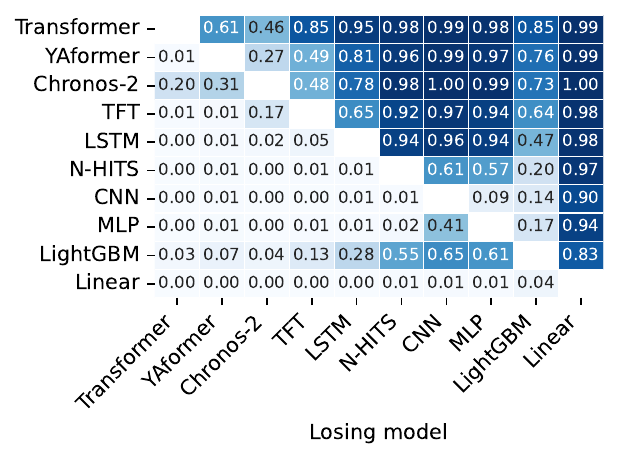}
        \caption{FeederBW}
        \label{fig:feederbw-results-statistical-test}
    \end{subfigure}
    \begin{subfigure}{0.33\textwidth}
        \includegraphics[width=\textwidth]{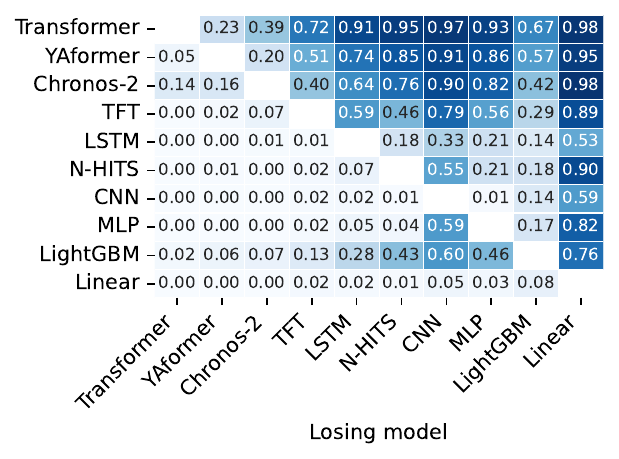}
        \caption{Electricity-287}
        \label{fig:electricity-results-statistical-test}
    \end{subfigure}
    \caption{Statistical test results on the three datasets. A Diebold-Mariano test is performed per time series for each pair of models. The fraction of time series on which the model in the column is better than the model in the row with $p<0.05$ is given.}
    \label{fig:statistical-tests}
 \end{figure*}

\begin{table*}
    \caption{Runtimes for one model training, one inference (one forecast for all time series), and annual total runtime (12 $\cdot$ training + 8760 $\cdot$ inference).}
    \label{tab:runtimes}
    \centering
    \begin{tabular}{lccccccccc}
    \toprule
    & \multicolumn{3}{c}{\textbf{TransnetBW}} & \multicolumn{3}{c}{\textbf{FeederBW}} & \multicolumn{3}{c}{\textbf{Electricity-287}} \\
    \cmidrule(lr){2-4} \cmidrule(lr){5-7} \cmidrule(lr){8-10}
     \textbf{Model} & Training & Inference & Annual & Training & Inference & Annual & Training & Inference & Annual \\
     & [h:mm:ss] & [ms] & [h:mm:ss] & [h:mm:ss] & [ms] & [hh:mm:ss] & [hh:mm:ss] & [ms] & [hh:mm:ss] \\
    \midrule
    Baseline    & - & 1.4 & 0:00:12 & - & 266.25 & 0:38:52 & - & 373.4 & 0:54:30 \\
    Chronos-2   & - & 78.3 & 0:11:26 & - & 11,160.7 & 27:09:27 & - & 16,015.6 & 38:58:16 \\
    CNN         & 0:03:53 & 5.4 & 0:47:29 & 0:36:06 & 28.1 & 7:17:19 & 1:42:37 & 35.8 & 20:36:37 \\
    LightGBM    & 0:18:42 & 1.2 & 3:44:36 & 10:04:12 & 269.7 & 121:29:46 & 50:20:28 & 239.3 & 604:40:32 \\
    Linear      & 0:05:59 & 5.2 & 1:12:39 & 0:37:22 & 38.5 & 7:17:19 & 1:11:35 & 38.9 & 14:28:16 \\
    LSTM        & 0:10:07 & 15.3 & 2:03:37 & 0:49:03 & 74.6 & 9:59:32 & 2:40:34 & 55.2 & 32:14:51 \\
    MLP         & 0:05:29 & 5.2 & 1:06:29 & 0:47:05 & 38.5 & 9:30:42 & 2:48:51 & 29.2 & 33:50:27 \\
    N-HITS      & 0:06:45 & 5.9 & 1:21:49 & 0:48:18 & 39.2 & 9:45:20 & 3:09:58 & 41.5 & 38:05:39 \\
    TFT         & 0:14:08 & 10.8 & 2:51:12 & 1:24:45 & 46.1 & 17:03:44 & 5:17:30 & 75.2 & 63:40:58 \\
    Transformer & 0:11:40 & 6.4 & 2:21:00 & 1:34:26 & 72.5 & 19:03:48 & 5:03:10 & 60.6 & 60:46:50 \\
    YAformer    & 0:11:33 & 6.3 & 2:19:32 & 0:50:25 & 82.6 & 10:17:04 & 5:07:09 & 95.2 & 61:39:42 \\
    \bottomrule
    \end{tabular}
\end{table*}

 \begin{figure*}
    \centering
    \begin{subfigure}{0.33\textwidth}
        \includegraphics[width=\textwidth]{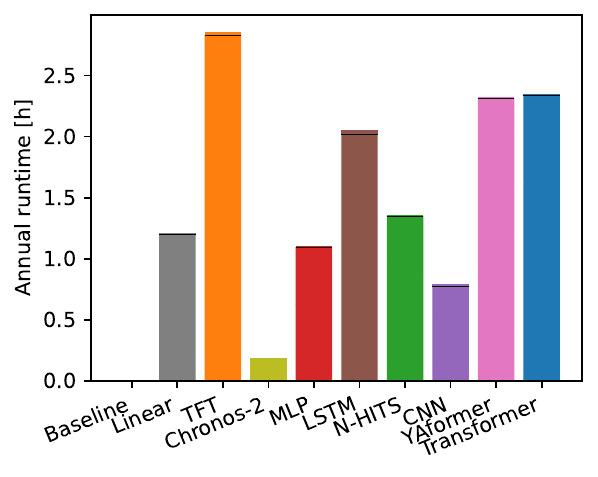}
        \caption{TransnetBW}
        \label{fig:transnetbw-runtimes}
    \end{subfigure}
    \begin{subfigure}{0.33\textwidth}
        \includegraphics[width=\textwidth]{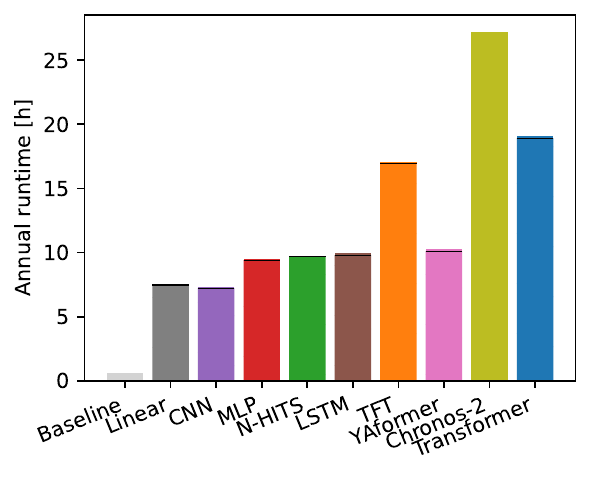}
        \caption{FeederBW}
        \label{fig:feederbw-runtimes}
    \end{subfigure}
    \begin{subfigure}{0.33\textwidth}
        \includegraphics[width=\textwidth]{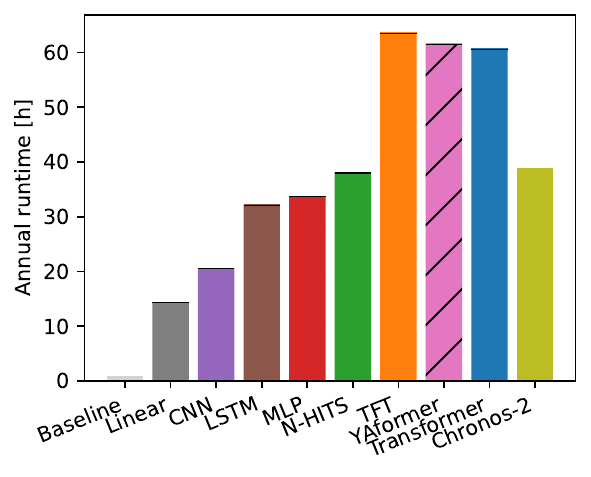}
        \caption{Electricity-287}
        \label{fig:electricity-runtimes}
    \end{subfigure}
    \caption{Annual total runtimes for twelve monthly model trainings and 8760 inferences on the three datasets. The training time is shown as dashed bar.}
    \label{fig:runtimes}
 \end{figure*}

\FloatBarrier

\subsection{Runtime}

Training and inference runtimes for all models are given in Table~\ref{tab:runtimes} (evaluated on a single \ac{HPC} node with an NVIDIA A100 GPU with \qty{40}{\giga\byte} VRAM, 19 CPU cores of an Intel Xeon Platinum 8368 and \qty{128}{\giga\byte} RAM).
The training time is the average time needed to train a global model.
The inference time is the average time needed to create a forecast for all time series in a dataset (one for TransnetBW, 200 for FeederBW, 287 for Electricity-287). The deep learning models trained by us are evaluated with a batch size equal to the number of time series, i.e. all time series are put into a single batch, whereas Chronos-2 is evaluated with a maximum batch size of 32, which resulted in the fastest throughput.
We calculate the annual runtime in a scenario where the models are retrained every month and a forecast is made at every hour for the next 96 hours.
This means that twelve models are trained and 8760 forecasts are made per year, so the annual runtime consists of twelve trainings and 8760 inference instances.
This annual runtime is visualized in Figure~\ref{fig:runtimes}, with the training part dashed.
For the models trained by us, the annual runtime is dominated by the training time, which comprises almost all of the annual runtime. The inference time is fast, ranging between \qty{5}{\milli\second} and \qty{95}{\milli\second}  per inference on the three datasets, so that it is barely visible in Figure~\ref{fig:results}.
Chronos-2 is used as a pretrained model, so it has zero training time in our scenario. This makes it the fastest model overall on the TransnetBW dataset, where only one time series needs to be forecast.
On the other datasets, which have 200 and 287 time series, the slower inference time of Chronos-2 outweighs the fact that it has zero training time.

\subsection{Ablations}

Figure~\ref{fig:ablations} shows results of ablated models, where covariates are removed, static training instead of monthly retraining is performed, and the context length is altered.

\paragraph{Covariates}
Covariates are important to achieve the lowest forecast errors. This holds for all three datasets, as the Transformer with all features is always better than the univariate Transformer.
The same holds for Chronos-2, which also benefits from covariates.
The improvement by the integration of covariates is large on the TransnetBW and FeederBW datasets, which are largely influenced by behind-the-meter photovoltaic power generation. The effect is smaller, but also important, on the Electricity-287 dataset, where some of the load is also driven by weather effects.
Apart from weather covariates, calendar features are important on the TransnetBW and Electricity-287 datasets, as the industrial activity is reduced during holidays, leading to a lower electrical load.



\paragraph{Continuous retraining}
On TransnetBW and FeederBW, a rolling evaluation is performed on a monthly basis, meaning that new models are trained every month. The previous month is used as validation data, and all data before that as training data, so that models can adapt to concept drifts due to newly installed appliances or changed consumption patterns.
We compare the retrained Transformer to a static model that is only trained once per year.
The retrained Transformer outperforms the static counterpart on both datasets, signifying that retraining helps the model to adapt.

\paragraph{Context length}
Figures~\ref{fig:ablations}d-f show the forecast error of Chronos-2 with one week, two weeks, one month context and the full context length of 8192 time steps.
The forecast accuracy of Chronos-2 improves with longer contexts on all three datasets.
With only one week context, Chronos-2 barely outperforms the baseline, but it quickly improves with longer contexts.
Using the maximum context of 8192 time steps gives only a small improvement over using one month context.

 \begin{figure*}
    \centering
    \begin{subfigure}{0.33\textwidth}
        \includegraphics[width=\textwidth]{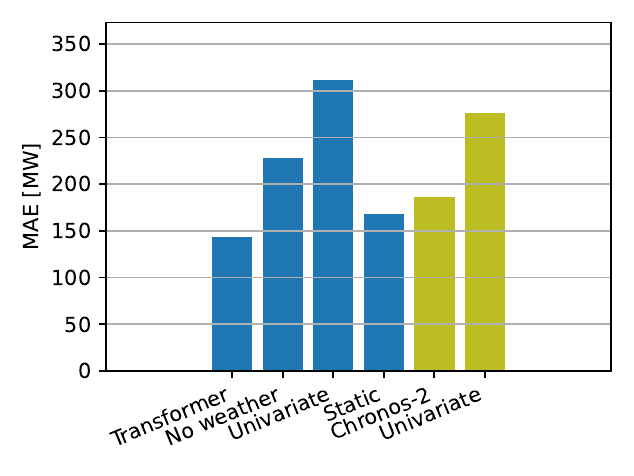}
        \caption{Ablations on TransnetBW}
        \label{fig:transnetbw-ablations}
    \end{subfigure}
    \begin{subfigure}{0.33\textwidth}
        \includegraphics[width=\textwidth]{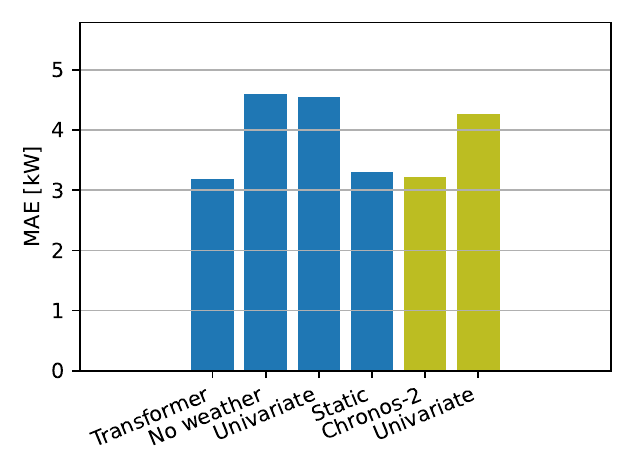}
        \caption{Ablations on FeederBW}
        \label{fig:transnetbw-ablations}
    \end{subfigure}
    \begin{subfigure}{0.33\textwidth}
        \includegraphics[width=\textwidth]{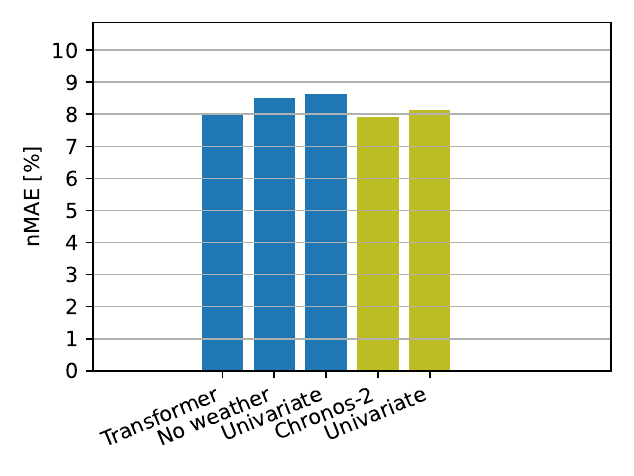}
        \caption{Ablations on Electricity-287}
        \label{fig:transnetbw-ablations}
    \end{subfigure}
    \begin{subfigure}{0.33\textwidth}
        \includegraphics[width=\textwidth]{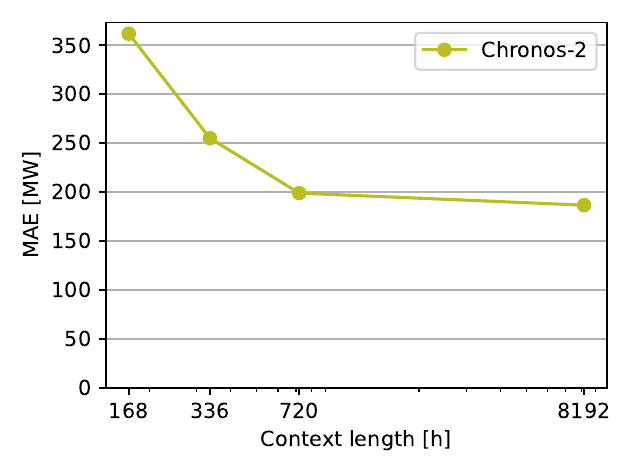}
        \caption{Context length on TransnetBW}
        \label{fig:transnetbw-context}
    \end{subfigure}
    \begin{subfigure}{0.33\textwidth}
        \includegraphics[width=\textwidth]{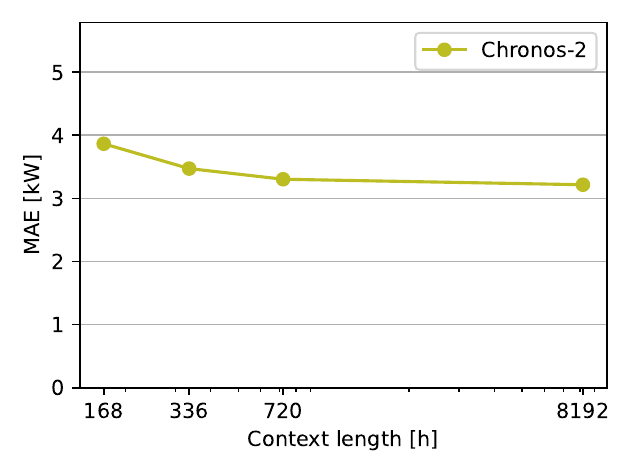}
        \caption{Context length on FeederBW}
        \label{fig:feederbw-context}
    \end{subfigure}
    \begin{subfigure}{0.33\textwidth}
        \includegraphics[width=\textwidth]{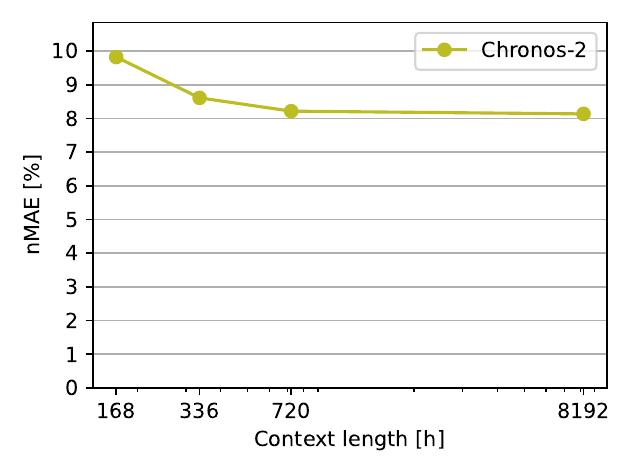}
        \caption{Context length on Electricity-287}
        \label{fig:electricity-context}
    \end{subfigure}
    \caption{Ablation results on test data for models without weather data (\emph{no weather}), without covariates (\emph{univariate}), and without retraining (\emph{static}), as well as reduced context length for Chronos-2.}
    \label{fig:ablations}
 \end{figure*}

\section{Discussion}
\label{sec:discussion}


Transformer-based models are effective for load forecasting on all tested grid levels. On the three datasets, the improvement over the best non-Transformer model is considerable, with 6.6--10.7\,\% lower errors for the best Transformer compared to the best other method. This includes datasets with different characteristics, such as highly aggregated TSO load and more diverse and volatile \ac{LV} feeder loads and client loads, where time series from multiple hundred measurement points are available. After hyperparameter optimization, Transformer models are effective for all these data types.

The architectural modifications to the Transformer -- including patching, \ac{CNN} layers, \ac{LSTM} layers, max pooling and sparse attention -- are not successful in reducing the forecast error. Instead, the standard Transformer outperforms the flexible \ac{YAformer} after hyperparameter optimization by a small margin.
This finding is further strengthened by the fact that the standard Transformer also outperforms the established Transformer-based model \ac{TFT} in our experiments.
Although the standard Transformer is a subclass of \ac{YAformer}, the larger hyperparameter search space of \ac{YAformer} hinders the \ac{HPO} in finding the optimal configuration. In fact, there is no guarantee that the Transformer configuration (or the configuration of any other model) after \ac{HPO} is the global optimum, but we had to restrict the \ac{HPO} to 100 evaluations per model for practical reasons.

Chronos-2 delivers accurate zero-shot forecasts for diverse time series, but it has particular weaknesses with rare events and long horizons, which lead to larger overall errors on the \ac{TSO} data. While Chronos-2 is very accurate for short forecast horizons between a few hours up to one day, it is outperformed by models trained particularly for the datasets at hand for longer horizons. On the TransnetBW data, it exhibits large errors around holidays, where it needs to incorporate not only the lower industrial activity on the holiday, but also indirect effects such as long weekends and school holidays. Future work can try to improve the results from Chronos-2 by designing additional features or by fine-tuning the model.

The advantage of Chronos-2 is that no dataset-specific training is required. Training a model from scratch takes between half an hour and five hours, depending on the dataset and model. This computational cost is saved with Chronos-2. On the other hand, Chronos-2 is a very large model, resulting in slower inference times. Inference times for our models are below \qty{100}{\milli\second} for creating one forecast for all time series in a dataset, whereas Chronos-2 needs up to 21 seconds. Therefore, Chronos-2 is useful when inference time plays a minor role, e.g. because few time series need to be forecasted or forecasts are only made once per day. For continuous forecasts during the day and for scenarios with many time series, dataset-specific models can be advantageous.

Long contexts, covariates and continuous retraining are important to achieve the lowest electrical load forecast errors.
This means that univariate \acp{TSFM} will likely not succeed in achieving the lowest possible forecast errors.
Comparisons on datasets without covariates, like the original UCI Electrical Load Diagrams, can indicate which models capture autocorrelated energy demand patterns, but are to be taken with caution because they do not show the full picture.

Deterministic forecasts do not cover the uncertainty of a forecast. This is especially relevant for volatile time series like FeederBW and Electricity-287. On both datasets, it is difficult to make good point forecasts due to the volatile and unpredictable pattern of the load, so that even the best models sometimes exhibit large errors. In such cases, it can be beneficial to create probabilistic forecasts that cover the uncertainty in form of predicted distributions. Future work can extend our benchmarking to probabilistic forecasting models. Chronos-2 offers the capability to predict quantiles. The deep learning models from our work can be extended to probabilistic models by replacing the linear output layer with a probabilistic output layer and training the models with the pinball loss function.

We agree with scholars~\cite{haben_review_2021, hewamalage_forecast_2023} that benchmarking of electrical load forecasting methods on different public datasets, with comparable test splits and metrics, is necessary to establish the state of the art in electrical load forecasting.
Therefore, we provide a comparison of a large set of methods on three datasets. We share our preprocessed datasets publicly, together with the predictions of all evaluated models, to allow for future comparisons to our results\footnote{The datasets and results are available on GitHub: \url{https://github.com/KIT-IAI/load-forecasting-benchmark}}.

The achieved forecast errors might be unrealistically low for real applications. For example, real-time measurements of past load values might become available only after a few hours or days, but are assumed to be available immediately in our experiments. In addition, we have used weather data that is not available at the forecast time in practice, such as reanalysis data or short-term weather forecasts. However, our primary goal is to compare models under fair conditions, which we achieve by using the same data for all.

\section{Conclusion \& Outlook}
\label{sec:conclusion}

The present paper presents a benchmark for electrical load forecasting across grid levels, including three datasets representing the \ac{TSO} level, the \ac{LV} grid feeder level and the end consumer level.
In a comparison of time-series Transformers and the \ac{TSFM} Chronos-2 with established machine learning and deep learning methods for electrical load forecasting on the three datasets, Transformer-based models outperform the established methods across all grid levels, reducing the forecast error by 6.6--10.7\,\%.
The standard time-series Transformer outperforms \ac{YAformer}, a flexible Transformer with patching, \ac{CNN} layers, \ac{LSTM} layers and sparse attention. This indicates that the standard architecture is already a strong model, and architectural modifications are not needed and can even lead to worse accuracy as it makes \acf{HPO} more difficult due to an extended hyperparameter search space.
The \ac{TSFM} Chronos-2 gives competitive forecasts on the feeder and client levels, being particularly accurate for short forecast horizons, but it is outperformed by specialized models on the \ac{TSO} level and it is comparably slow in inference.

All our datasets and results are made publicly available to enable benchmarking of new approaches against our results, as well as to allow for an extension to probabilistic models in the future.
\ifanonymize{We plan to evaluate the best models continuously using a public platform with real-time \ac{TSO} data for model comparisons under realistic operational conditions~\cite{kleinebrahm_energy-arena_2026_anonymized}.}{We plan to integrate the best approaches into the Energy Arena\footnote{\url{https://energy-arena.org}}~\cite{kleinebrahm_energy-arena_2026} for continuous performance comparisons under realistic operational conditions.}
Ensembling of forecast models could improve the results beyond single-model forecasts.

\begin{acks}
The authors thank Alexander Kreusel for assistance with Chronos-2.

The authors gratefully acknowledge funding by the Helmholtz Association under the program “Energy System Design” and the Helmholtz Association’s Initiative and Networking Fund through Helmholtz AI and the HAICORE@KIT partition.
The authors gratefully acknowledge the computing time provided on HoreKa by the NHR@KIT center. This center is jointly supported by the Federal Ministry of Education and Research and the Ministry of Science, Research and the Arts of Baden-Württemberg, as part of the National High-Performance Computing (NHR) joint funding program.
HoreKa is partly funded by the German Research Foundation (DFG).

ChatGPT was used to assist in parts of the writing and coding. All outputs were edited carefully and the authors take the full responsibility for the published article.
\end{acks}


\bibliographystyle{ACM-Reference-Format}
\bibliography{zotero-references,additional-references}

\appendix

\begin{table*}
    \centering
    \caption{Number of trainable parameters per model on the three datasets.}
    \label{tab:model-sizes}
    \begin{tabular}{lrrr}
    \toprule
    Model & \multicolumn{1}{c}{TransnetBW} & \multicolumn{1}{c}{FeederBW} & \multicolumn{1}{c}{Electricity-287} \\
    \midrule
    \ac{CNN}      &  8,263,904 &   7,710,304 &   3,692,768 \\
    Linear        &    940,128 &     456,288 &     940,128\\
    \ac{LSTM}     & 13,855,456 & 117,325,408 & 117,327,456 \\
    \ac{MLP}      &  2,966,624 &  14,127,200 &   6,457,440 \\
    N-HITS        &  7,308,894 &  48,654,500 &  24,865,431 \\
    \ac{TFT}      &    195,828 &     696,810 &     741,748 \\
    Transformer   &  1,078,657 &   3,304,961 &     279,361 \\
    \ac{YAformer} &  4,350,849 &   2,823,297 &   8,608,769 \\
    \bottomrule
    \end{tabular}
\end{table*}

\newcommand{\bb}[1]{\bfseries #1}

\begin{table*}
    \caption{Test results on the three datasets evaluated with the \acf{MAE}, the \acf{RMSE}, the \acf{nMAE} and the \acf{nRMSE}. The best result is highlighted in bold.}
    \label{tab:metric-results}
    \centering
    \begin{tabular}{lSSSSSSSSSSSS}
    \toprule
     \textbf{Model} & \multicolumn{4}{c}{\textbf{TransnetBW}} & \multicolumn{4}{c}{\textbf{FeederBW}} & \multicolumn{4}{c}{\textbf{Electricity-287}} \\
     \cmidrule(lr){2-5} \cmidrule(lr){6-9} \cmidrule(lr){10-13}
      & \ac{MAE} & \ac{RMSE} & \ac{nMAE} & \ac{nRMSE} & \ac{MAE} & \ac{RMSE} & \ac{nMAE} & \ac{nRMSE} & \ac{MAE} & \ac{RMSE} & \ac{nMAE} & \ac{nRMSE} \\
      & [MW] & [MW] & [\%] & [\%] & [kW] & [kW] & [\%] & [\%] & [kW] & [kW] & [\%] & [\%] \\
    \midrule
    Baseline    & 355.3 & 487.1 & 5.49 & 7.53 & 5.51 & 9.24 & 33.5 & 51.8 & 56.1 & 566.5 & 10.35 & 16.79 \\
    Chronos-2   & 186.5 & 280.0 & 2.88 & 4.33 & 3.22 & 4.91 & 19.4 & 28.4 & 42.2 & 407.0 &  \bb{7.92} & 12.95 \\
    CNN         & 156.5 & 211.1 & 2.42 & 3.26 & 3.66 & 5.41 & 22.2 & 31.3 & 47.3 & 430.2 &  8.97 & 13.34 \\
    LightGBM    & 186.0 & 259.8 & 2.87 & 4.01 & 3.49 & 5.31 & 21.1 & 30.6 & 43.6 & 377.3 &  8.86 & 13.08 \\
    Linear      & 237.7 & 322.2 & 3.67 & 4.98 & 3.99 & 6.00 & 24.2 & 34.4 & 54.1 & 508.7 &  9.76 & 14.08 \\
    LSTM        & 163.8 & 217.5 & 2.53 & 3.36 & 3.41 & 5.05 & 20.6 & 29.2 & 48.2 & 453.1 &  9.10 & 13.77 \\
    MLP         & 166.0 & 222.4 & 2.57 & 3.44 & 3.64 & 5.40 & 22.0 & 31.1 & 47.9 & 438.3 &  8.98 & 13.25 \\
    NHITS       & 160.5 & 215.7 & 2.48 & 3.33 & 3.60 & 5.32 & 21.8 & 30.7 & 47.5 & 445.3 &  8.92 & 13.33 \\
    TFT         & 191.3 & 259.5 & 2.96 & 4.01 & 3.30 & 5.01 & 20.0 & 28.9 & 44.1 & 413.3 &  8.51 & 13.47 \\
    Transformer & \bb{143.6} & \bb{191.7} & \bb{2.22} & \bb{2.96} & \bb{3.18} & \bb{4.82} & \bb{19.2} & \bb{27.8} & \bb{40.6} & \bb{378.9} &  7.99 & \bb{12.56} \\
    YAformer    & 148.4 & 199.8 & 2.29 & 3.09 & 3.24 & 4.90 & 19.6 & 28.4 & 42.7 & 415.1 &  8.18 & 12.77 \\
    \bottomrule
    \end{tabular}
\end{table*}

\section{Hyperparameters}
\label{sec:hyperparameters}

The hyperparameters of the best configurations found with the \ac{HPO} are given in Table~\ref{tab:hyperparameters} and the resulting model sizes in Table~\ref{tab:model-sizes}.

\section{Additional metrics}
\label{sec:metric-results}

The \acf{MAE}, \acf{RMSE}, \acf{nMAE} and \acf{nRMSE} results on the test sets of the three datasets are given in Table~\ref{tab:metric-results}.

\begin{table*}
    \scriptsize
    \centering
    \caption{Model hyperparameter search range and the best configurations found with a Bayesian Optimization \ac{HPO} for each dataset.}
    \label{tab:hyperparameters}
    \vspace{-0.25cm}
    \begin{tabular}{lcccc}
    \toprule
    Hyperparameter & Search range & TransnetBW & FeederBW & Electricity-287 \\
    \midrule
     \underline{\ac{CNN}} \\
    ~~Context length $C$ & 168, 336, 720 & 336 & 168 & 336 \\
    ~~Patch size & 1, 4, 8, 24 & 8 & 24 & 24 \\
    ~~Kernel width & 3, 5, 7 & 3 & 7 & 5 \\
    ~~Convolutional layers & 1, 2, 3 & 1 & 1 & 1 \\
    ~~Max pooling width & 1, 2, 4, 8 & 1 & 1 & 1 \\
    ~~Dense layers & 1, 2, 3 & 2 & 1 & 2 \\
    ~~Dense units & 128, 256, 512, 1024 & 1024 & 512 & 1024 \\
    ~~Batch size & 128, 256, 512 & 256 & 128 & 512 \\
    ~~Learning rate & (0.0001, 0.001) & 0.00079 & 0.00016 & 0.00014 \\
    ~~Dropout & (0.0, 0.2) & 0.00 & 0.09 & 0.12 \\
    \underline{LightGBM} \\
    ~~Context length $C$ & 168 & 168 & 168 & 168 \\
    ~~Learning rate & (0.01, 0.3) & 0.153 & 0.0492 & 0.0519 \\
    ~~Minimum child samples & (1, 200) & 32 & 58 & 7 \\
    ~~Number of estimators & (1, 250) & 248 & 232 & 211 \\
    \underline{Linear} \\
    ~~Context length $C$ & 168, 336, 720 & 720 & 336 & 720 \\
    ~~Batch size & 128, 256, 512 & 512 & 512 & 512 \\
    ~~Learning rate & (0.0001, 0.001) & 0.00012 & 0.00010 & 0.00017 \\
    ~~Dropout & (0.0, 0.2) & 0.15 & 0.08 & 0.05 \\
    \underline{\ac{LSTM}} \\
    ~~Context length $C$ & 168, 336, 720 & 720 & 720 & 720 \\
    ~~Architecture & encoder, encoder-decoder & encoder & encoder & encoder \\
    ~~Patch size & 1, 4, 8, 24 & 4 & 4 & 4 \\
    ~~LSTM layers & 1, 2, 3 & 2 & 3 & 3 \\
    ~~$d_\mathrm{model}$ & 32, 64, ..., 1024 & 128 & 512 & 512 \\
    ~~Dense units & 128, 256, 512, 1024 & 512 & 1024 & 1024 \\
    ~~Dense layers & 1, 2, 3 & 1 & 1 & 1 \\
    ~~Batch size & 128, 256, 512 & 256 & 128 & 128 \\
    ~~Learning rate & (0.0001, 0.001) & 0.00080 & 0.00037 & 0.00012 \\
    ~~Dropout & (0.0, 0.2) & 0.19 & 0.04 & 0.08 \\
    \underline{\ac{MLP}} \\
    ~~Context length $C$ & 168, 336, 720 & 336 & 336 & 336 \\
    ~~Layers & 1, 2, 3 & 2 & 2 & 2 \\
    ~~Units & 128, 256, 512, 1024, 2048 & 512 & 2048 & 1024 \\
    ~~Batch size & 128, 256, 512 & 512 & 256 & 512 \\
    ~~Learning rate & (0.0001, 0.001) & 0.00019 & 0.00018 & 0.00012 \\
    ~~Dropout & (0.0, 0.2) & 0.00 & 0.12 & 0.09 \\
    \underline{N-HITS} \\
    ~~Context length $C$ & 168, 336, 720 & 168 & 336 & 720 \\
    ~~Blocks & (1,1,1), (2,2,2), (3,3,3) & (1,1,1) & (2,2,2) & (3,3,3) \\
    ~~Kernel size & (1,1,1), (2,2,2), (4,4,4), (2,2,1), (8,4,1), (16,8,1) & (2,2,1) & (8,4,1) & (16,8,1) \\
    ~~Downsample & (1,1,1), (24,12,1), (40,20,1), (60,8,1), (168,24,1), (180,60,1) & (40,20,1) & (180,60,1) & (60,8,1) \\
    ~~Dense units & 32, 64, 128, ..., 1024 & 512 & 1024 & 256 \\
    ~~Batch size & 128, 256, 512 & 256 & 512 & 128 \\
    ~~Learning rate & (0.0001, 0.001) & 0.00025 & 0.00014 & 0.00010 \\
    \underline{\ac{TFT}} \\
    ~~Context length $C$ & 168, 336, 720 & 720 & 336 & 720 \\
    ~~$d_\mathrm{model}$ & 32, 64, 128, 256, 512 & 32 & 64 & 64 \\
    ~~$n_\mathrm{heads}$ & 1, 2, 4, 8 & 1 & 4 & 4 \\
    ~~Batch size & 128, 256, 512 & 128 & 128 & 128 \\
    ~~Learning rate & (0.0001, 0.001) & 0.00049 & 0.00010 & 0.00035 \\
    ~~Dropout & (0.0, 0.2) & 0.04 & 0.03 & 0.02 \\
    \underline{Transformer} \\
    ~~Context length $C$ & 168, 336, 720 & 720 & 720 & 720 \\
    ~~$n_\mathrm{layers}$ & 1, 2, 3, 4 & 1 & 3 & 4 \\
    ~~$d_\mathrm{model}$ & 32, 64, 128, 256, 512 & 256 & 256 & 64\\
    ~~$n_\mathrm{heads}$ & 1, 2, 4, 8 & 1 & 4 & 4\\
    ~~Dense layers & 1, 2, 3 & 1 & 2 & 1 \\
    ~~Dense units & 64, 128, 256, 512 & 64 & 256 & 128 \\
    ~~Batch size & 128, 256, 512 & 128 & 512 & 512 \\
    ~~Learning rate & (0.0001, 0.001) & 0.00028 & 0.00017 & 0.00033 \\
    ~~Dropout & (0.0, 0.2) & 0.17 & 0.10 & 0.07 \\
    \underline{\ac{YAformer}} \\
    ~~Context length $C$ & 168, 336, 720 & 336 & 720 & 720 \\
    ~~Patch size & 1, 4, 8, 24 & 1 & 1 & 1 \\
    ~~$w_\mathrm{conv}$ & 1, 3, 5 & 5 & 1 & 5\\
    ~~LSTM layers & 0, 1, 2 & 0 & 1 & 0 \\
    ~~$n_\mathrm{layers}$ & 1, 2, 3, 4 & 1 & 1 & 2 \\
    ~~$d_\mathrm{model}$ & 32, 64, 128, 256, 512 & 512 & 256 & 512 \\
    ~~$n_\mathrm{heads}$ & 1, 2, 4, 8 & 2 & 1 & 8 \\
    ~~Attention & full, sparse & full & sparse & full \\
    ~~Max pooling width & 1, 2, 4 & 4 & 1 & 1 \\
    ~~Dense layers & 1, 2, 3 & 2 & 2 & 1 \\
    ~~Dense units & 64, 128, 256, 512 & 128 & 128 & 256\\
    ~~Batch size & 128, 256, 512 & 256 & 128 & 512 \\
    ~~Learning rate & (0.0001, 0.001) & 0.00014 & 0.00038 & 0.00056 \\
    ~~Dropout & (0.0, 0.2) & 0.05 & 0.11 & 0.10 \\
    \bottomrule
    \end{tabular}
\end{table*}

\end{document}
\endinput